\documentclass[11pt,a4paper]{article}
\usepackage[T1]{fontenc}
\usepackage[utf8]{inputenc}
\usepackage{lmodern}
\usepackage[margin=1in]{geometry}
\usepackage{amsmath,amssymb}
\usepackage{microtype}
\usepackage{graphicx}
\usepackage{url}
\usepackage[hidelinks]{hyperref}
\newcommand{\DLH}{\mathcal H}
\newcommand{\DLN}{\mathcal N}
\newcommand{\DLC}{\mathcal C}
\newcommand{\DLE}{\mathbb E}
\newcommand{\DLR}{\mathbb R}
\newcommand{\DLnorm}[1]{\lVert #1\rVert}

\newenvironment{acknowledgement}{\par\medskip\noindent\begin{minipage}{\linewidth}\small\textbf{Acknowledgements}\quad\ignorespaces}{\par\end{minipage}\par}

\title{Accelerating Posterior Sampling in Bayesian Inverse Problems with Diffusion Models: An Online--Offline Perspective}
\author{Duc-Lam Duong\\[0.5em]
{\small Research Unit of Mathematical Sciences, University of Oulu,}\\
{\small P.O. Box 8000, FI-90014 University of Oulu, Finland}\\
{\small\texttt{duc-lam.duong@oulu.fi}}}
\date{}
\hypersetup{
  pdftitle={Accelerating Posterior Sampling in Bayesian Inverse Problems with Diffusion Models: An Online--Offline Perspective},
  pdfauthor={Duc-Lam Duong}
}

\begin{document}
\maketitle

\begin{abstract}
Bayesian inverse problems require both informative prior models and computationally feasible posterior inference. Score-based diffusion models provide expressive learned priors, but their use for inversion introduces a conditioning problem: the score of a noised posterior is generally unavailable from a prior score alone. This selective survey reviews the principal approaches to that problem, including likelihood guidance, spectral and projection methods, sequential Monte Carlo corrections, and amortized conditional training. We organize methods by how they split cost between offline preparation and online inference, and compare their assumptions, advantages and limitations, with emphasis on computational speedups. 
Some ongoing directions and open questions are gathered at the end.
\end{abstract}

\noindent\textbf{Key words:} Bayesian inverse problems; score-based diffusion models; posterior sampling; amortized inference; uncertainty quantification\par
\smallskip
\noindent\textbf{MSC2020:} 62F15; 65C40; 65N21; 68T07.

\section{Introduction}
\label{sec:intro}

\subsection{Infinite-dimensional Bayesian inverse problems}
Inverse problems deal with the task of recovering a state, coefficient or source from indirect observations. Examples include reconstructing internal body structures from X-ray measurements  in CT imaging and identifying coefficients from observations of the solution of a partial differential equation (PDE). Due to the \emph{ill-posed} nature (nonexistence, nonuniqueness and instability) of inverse problems, one often requires additional information.
Bayesian inversion represents this information by a \emph{prior} measure and conditions it on observations to obtain a \emph{posterior} \cite{DL_Stuart2010}. Here we adopt an infinite-dimensional framework that allows the unknown to live in, say, a Hilbert space. This permits well-posedness and approximation to be investigated before discretization, which encourages developments of discretization-invariant algorithms whose convergence can be studied under mesh refinement. 

Let $X$ take values in a separable Hilbert space $\DLH$, with prior $\mu$, and
consider the observation model
\begin{equation}\label{eq:obs}
 Y=\mathcal G(X)+\varepsilon,\qquad
 \varepsilon\sim\DLN(0,\Gamma),
\end{equation}
where $\mathcal G:\DLH\to\DLR^m$ is measurable, $\Gamma$ is positive definite, and $\varepsilon$ is independent of $X$. The posterior $\mu^y$ satisfies
\begin{equation*}\label{eq:bayes}
 \frac{\mathrm d\mu^y}{\mathrm d\mu}(x)
 =\frac{\exp[-\Phi(x;y)]}{Z(y)},\qquad
 \Phi(x;y)=\frac12\DLnorm{\mathcal G(x)-y}_{\Gamma}^{2},
\end{equation*}
with $\DLnorm{v}_{\Gamma}^{2}=v^\top\Gamma^{-1}v$ and $Z(y)=\int\exp[-\Phi(x;y)]\,\mu(\mathrm dx)>0$.
The posterior carries information about the unknown and its  uncertainty. 

Posterior computation can nevertheless be expensive. Markov chain Monte Carlo
(MCMC) provides asymptotically exact inference under suitable conditions, and
function-space methods avoid some deterioration under mesh refinement
\cite{DL_Cotter2013}. Slow exploration and costly likelihood evaluations still
limit effective sample size, especially when $\mathcal G$ requires PDE solves.
Surrogates and transport maps offer alternative approaches \cite{DL_Cui2015, DL_Marzouk2017}.

Denoising diffusion probabilistic models \cite{DL_Ho2020} and the continuous score-based formulation \cite{DL_Song2021} learn noised distributions and generate samples by reversing the noising process. Inversion requires incorporating the likelihood. Early medical-imaging work demonstrated reusable unconditional scores
\cite{DL_SongMed2022}, and subsequent conditioning methods have different accuracy and cost properties. See the review papers \cite{DL_Daras2024,DL_Janati2025} and upcoming book \cite{Lai2027principles} for recent developments.


\subsection{Diffusion priors and posterior conditioning}
\label{sec:diffusion}

Consider the Ornstein--Uhlenbeck (OU) process
\begin{equation}\label{eq:ou}
 \mathrm dX_t=-\frac12X_t\,\mathrm dt+\mathrm dW_t,
 \qquad X_0\sim\nu,
\end{equation}
where $W$ is a $\DLC$-Wiener process independent of $X_0$, and $\DLC$ is an
injective, positive, self-adjoint, trace-class operator. Then
$X_t=e^{-t/2}X_0+\sqrt{1-e^{-t}}Z$ in distribution, with
$Z\sim\DLN(0,\DLC)$ independent of $X_0$. For a law $\nu$ with finite second
moment, define the covariance-weighted score (\cite{DL_Pidstrigach2024})
\begin{equation}\label{eq:score}
 s(x,t;\nu)=-\frac{x-e^{-t/2}\DLE[X_0\mid X_t=x]}{1-e^{-t}}.
\end{equation}
In finite dimensions this equals $\DLC\nabla\log p_t(x)$, where $p_t$ is the
density of $X_t$. Denoising score matching learns this field without needing an explicit density for $\nu$ \cite{DL_Pidstrigach2024}.

In function space there is no Lebesgue density, and \eqref{eq:score} provides a useful probabilistic definition. 
Under appropriate support and integrability assumptions, time reversal with a $\DLC$-Wiener process $\overline W$ takes the form
\begin{equation}\label{eq:reverse}
 \mathrm dZ_u=\left[\frac12Z_u+s(Z_u,T-u;\nu)\right]\mathrm du
 +\mathrm d\overline W_u,\qquad 0\leq u<T.
\end{equation}
With the exact terminal law as initialization, this reverses \eqref{eq:ou}. See \cite{DL_Pidstrigach2024} for detailed derivations and proofs.

For posterior sampling, the required initial law is $\nu=\mu^y$. In finite
dimensions, let $p_t$ denote the noised prior density and $p_t^y$ the noised
posterior density. Bayes' formula gives
\begin{equation}\label{eq:guidance}
 \nabla\log p_t^y(x)=\nabla\log p_t(x)+\nabla\log h_t(x;y),\quad
 h_t(x;y)=\DLE[e^{-\Phi(X_0;y)}\mid X_t=x].
\end{equation}
The expectation uses the prior noising process. The likelihood depends on $X_0$, so the second term in \eqref{eq:guidance} is usually unavailable even with a known likelihood and prior score. Existing works introduced different ways to estimate \eqref{eq:guidance}, leading to multiple posterior sampling strategies. 

\section{Approaches to efficient posterior sampling}
\label{sec:methods}

Below is a sample\footnote{This list is by no means exhaustive; the reader is referred to \cite{DL_Daras2024} for a broader review.} of different posterior sampling methods using diffusion models. 
These methods differ in how they incorporate observations and distribute computation between \emph{offline preparation} and \emph{online inference}. Offline work may include prior training, simulated observation pairs and operator-dependent preprocessing. Online work is performed for each new dataset and includes conditioning and sample generation. 

\subsection{Reusing pretrained prior models}
\label{subsec:guidance}

These methods reuse a pretrained prior model and incorporate the observation model during inference.
Diffusion posterior sampling (DPS) \cite{DL_Chung2023} approximates the
intermediate likelihood using a denoised estimate
$\widehat x_0(x,t)=\DLE[X_0\mid X_t=x]$. For Gaussian observations, the resulting
guidance is based on
$-\nabla_x\Phi(\widehat x_0(x,t);y)$. This accommodates differentiable nonlinear maps and, with other likelihoods,
non-Gaussian noise, without joint training on parameter--observation pairs.
However, guidance requires differentiating through the denoiser and forward model
and typically needs tuning. Evaluating a likelihood at the denoising mean remains
an approximation even with an exact prior score.

Tweedie moment projected diffusions (TMPD) \cite{DL_Boys2024} retain denoising
covariances as well as means. Moment-based approximations to $X_0\mid X_t=x$
produce tractable intermediate likelihoods for linear-Gaussian observations,
accounting for conditional spread at additional derivative or covariance cost. For non-Gaussian priors, the resulting moment approximation generally retains conditioning bias. Feynman--Kac analysis
\cite{DL_Delgadino2026} characterizes bias in DPS and related guidance even with
exact prior scores, and analyses strong-guidance discretization instabilities.

Denoising diffusion restoration models (DDRM) \cite{DL_Kawar2022} exploit linear
structure through noise-aware updates in singular-vector coordinates. They enable
efficient restoration with reusable priors and few denoising steps, but require
tractable spectral representations. Denoising diffusion null-space models (DDNM)
\cite{DL_Wang2023} use range--null-space decompositions to enforce data consistency
while the prior determines unobserved components. Both exploit operator structure, but reconstruction quality or noiseless
consistency does not establish calibrated posterior sampling.


The same pretrained priors can also be conditioned using sequential Monte Carlo (SMC), which approximates intermediate distributions with weighted particles, resampling and mutation. MCGDiff \cite{DL_Cardoso2024} constructs intermediate linear problems whose posteriors approach the target as diffusion noise decreases. Filtering posterior sampling \cite{DL_Dou2024} relates conditioning to Bayesian filtering and obtains asymptotically accurate inference via SMC. Decoupled diffusion SMC (DDSMC) \cite{DL_Kelvinius2025} allows larger generative updates and is asymptotically exact for linear-Gaussian observations.

Particle corrections offer a principled alternative to approximate guidance. 
Finite-particle weight degeneracy can impair accuracy, while particles increase score evaluations, likelihood computations and memory. Their common use of twisting intermediate distributions is reviewed
in \cite{DL_Janati2025}.

\subsection{Learning posterior conditioning}
\label{subsec:conditional}

Conditional training moves the incorporation of observations into
the learning stage. A network $s_\theta(x,t,y)$ trained on joint
samples of $(X,Y)$ learns the noised posterior score and can generate
samples without repeated forward or adjoint evaluations.
Baldassari et al. \cite{DL_Baldassari2024} justify conditional
denoising estimation for infinite-dimensional linear inverse
problems, including its small-time behaviour.
The online savings require representative joint training data.
Reuse across different operators, noise statistics or sensor
geometries requires those settings to be covered by training
or addressed through adaptation.

A related strategy learns the relationship between PDE parameters
and solutions. The main FunDPS implementation \cite{DL_Yao2025}
trains a joint diffusion model of these functions and incorporates
sparse observations through masking and gradient guidance.
Its alternative formulation uses a prior on the parameters alone
and an explicit PDE solver, retaining differentiation through
that solver during sampling. These formulations illustrate how
learning the forward relationship changes the balance between
offline simulations and online computation.

Learning only the guidance term provides an intermediate approach.
Baker et al. \cite{DL_Baker2026} establish an infinite-dimensional
Doob $h$-transform under suitable Cameron--Martin support or
Gaussian absolute-continuity assumptions and propose supervised
guidance training. This permits reuse of an existing prior network
while training an additional model for conditioning.
Their likelihood-informed parametrization nevertheless retains
observation-model gradients during sampling, so the computational
benefit depends on which parts of conditioning are learned.

\subsection{Reducing reverse-time evaluations}
\label{subsec:fast}

A complementary way to reduce online cost is to use fewer score or denoiser evaluations. DDIM \cite{DL_DDIM2021} permits sampling with fewer denoising steps, while DPM-Solver \cite{DL_Lu2022} uses high-order diffusion ODE integration. Both can use existing trained models without retraining. Consistency models
\cite{DL_Consistency2023} support one- or few-step generation through training or distillation, thereby moving additional computational effort into offline preparation.

For posterior sampling, these savings must be assessed together with conditioning and numerical integration errors.
Posterior-DDIM \cite{DL_Jiao2026} uses noise-dependent updates in singular directions and proves posterior convergence under its assumptions. 
Extending such guarantees to nonlinear inverse problems and function-space formulations remains an important direction.

\section{Task-dependent amortization: UCoS and regularized variant}
\label{sec:ucos}

The unconditional representation of the conditional score (UCoS)
\cite{DL_Schneider2025} combines a measurement-independent denoising
formulation with posterior sampling that avoids repeated physics
evaluations. Its training is specific to the observation model,
and the resulting network can be reused across new datasets for
that model.

Assume $\mathcal G=A:\DLH\to\DLR^m$ is bounded linear, with Gaussian observation noise (the prior need not be Gaussian) as in \eqref{eq:obs}. 
For $t>0$, set $q_t=e^t-1$ and define
\begin{equation*}
    C_t=\Gamma+q_tA\DLC A^*,\qquad R_t=q_tI-q_t^2\DLC A^*C_t^{-1}A,\qquad \Sigma_t=R_t\DLC.
\end{equation*}
Let $\widetilde X_t=\widetilde X_0+\eta_t$, with
$\widetilde X_0\sim\mu$ and $\eta_t\sim\DLN(0,\Sigma_t)$
independent, and put
\begin{equation*}\label{eq:r}
 r(z,t)=\DLE[\widetilde X_0\mid\widetilde X_t=R_tz],
 \qquad
 \lambda(t)=(e^{t/2}-e^{-t/2})^{-1}.
\end{equation*}
Combining the Gaussian likelihood with the Gaussian noising kernel by ``completing the square" gives the identity
\begin{equation*}\label{eq:ucos}
 s(x,t;\mu^y)=\lambda(t)\left[
 r\bigl(\DLC A^*\Gamma^{-1}y+\lambda(t)x,t\bigr)
 -e^{t/2}x\right].
\end{equation*}
In function space, this holds almost surely under the
Cameron--Martin support assumption
$\mu(\DLC^{1/2}(\DLH))=1$ \cite{DL_Schneider2025}.
The identity represents the conditional score exactly, without
approximating the likelihood. 

A network approximating $r$ is trained on prior samples with
perturbations depending on $(A,\Gamma,\DLC)$, independently of
the realized observation. Training requires operator calculations,
but admits a matrix-free implementation without constructing
the full operator matrix or its SVD. After computing
$b_y=\DLC A^*\Gamma^{-1}y$ once, reverse sampling uses $b_y$,
scalar operations and the network, without further applications
of $A$ or $A^*$.
Thus, UCoS avoids the separate measurement-vector input of full
conditional training while removing repeated online physics
evaluations. Changes to the observation operator or noise model
generally require retraining or adaptation.

The regularized variant of UCoS in \cite{DL_Schneider2026} combines a learned conditional score with the score of a Gaussian reference posterior,
\begin{equation}\label{eq:blend}
 s_\alpha=(1-\alpha)s_{\rm learned}+\alpha s_{\rm Gaussian},
 \qquad 0\leq\alpha\leq1.
\end{equation}
This supplements learned information in severely ill-posed
settings, including limited-view diffuse optical tomography.

To interpret this regularization, consider exact component scores
in finite dimensions with $\DLC=I$. Let $P_t^*$ denote OU density evolution and $p_1,p_2$ the initial posterior densities, underlying $s_{\rm learned}$ and $s_{\rm Gaussian}$, respectively.
The ideal score blend equals
$\nabla\log[(P_t^*p_1)^{1-\alpha}(P_t^*p_2)^\alpha]$.
It generally differs from the score
$\nabla\log P_t^*(p_1^{1-\alpha}p_2^\alpha)$ obtained by noising the geometric mixture. Under regularity and local positivity assumptions \cite{DL_Schneider2026}, this difference is $O(t)$ on bounded regions as $t\downarrow0$.
The result provides a local small-time interpretation of the ideal score blend. Note, however, that it does not establish that the learned regularized sampler exactly samples the geometric mixture.

\section{Conclusion and outlook}
\label{sec:conclusion}

Reusable prior guidance is flexible but usually approximate. Particle corrections offer asymptotic consistency at additional cost. Conditional and task-dependent training amortize conditioning, while fast reverse solvers reduce network evaluations. Their relative merits depend on operator structure, training data, reuse and required posterior accuracy. 
As a simple illustration, let \(C_{\mathrm{off}}\) denote the reusable training and preparation cost, and \(C_{\mathrm{on}}\) the cost per new dataset at a prescribed posterior accuracy. Reusing this preparation for \(N\) datasets gives an average cost per inversion of approximately \(C_{\mathrm{off}}/N+C_{\mathrm{on}}\). If method \(A\) has a greater offline cost but lower online cost than \(B\), its total cost would be smaller when
\[
N> (C_{\mathrm{off}}^{A}-C_{\mathrm{off}}^{B})/(C_{\mathrm{on}}^{B}-C_{\mathrm{on}}^{A}).
\]

The preceding discussion motivates the following open questions.
\emph{Nonlinear and non-Gaussian conditioning.}
Likelihood guidance and learned Doob transforms provide approaches
to conditioning beyond the linear-Gaussian setting
\cite{DL_Chung2023,DL_Baker2026}.
Can intermediate likelihoods or guidance terms be approximated
with controlled bias while reducing PDE and adjoint solves?
Of particular interest are guarantees that remain useful for multimodal or strongly concentrated posteriors.
Recent work \cite{Duong2026lwr} combines a diffusion prior with derivative-free likelihood sampling for initial identification in the LWR traffic model, with possible extensions to trajectory-based forward models \cite{dashti2023stability, duong2021inverse}. 
Further extensions to other nonlinear models require new ideas.

\emph{Learning and model error.}
Recent analysis quantifies how generative-prior approximation affects the posterior \cite{DL_Hosseini2026}.
Can these results be combined with finite-data score estimation, conditioning and numerical errors to obtain quantitative guarantees for posterior quantities of interest?
Understanding the dependence of these bounds on observation noise
and model misspecification is particularly important.
The statistically justified selection of score regularization
is a related problem.

\emph{Discretization and observations.}
Function-space diffusion theory provides dimension-independent
bounds under suitable assumptions
\cite{DL_Pidstrigach2024}, while conditional formulations reveal
additional difficulties near zero diffusion time
\cite{DL_Baldassari2024}.
Can conditional-score and numerical bounds be obtained uniformly under mesh refinement, varying sensors and small observation noise?
Moreover, the geometric interpretation of the mixed score \eqref{eq:blend} is restricted to finite dimensions and small diffusion times. Extending this interpretation to function space requires suitable absolute continuity, integrability and score regularity assumptions.

\emph{Posterior-preserving acceleration.}
Higher-order diffusion solvers and consistency models reduce
generation costs \cite{DL_Lu2022,DL_Consistency2023}.
Can analogous acceleration retain quantified posterior accuracy
after conditioning? Posterior convergence results for particular
linear constructions \cite{DL_Jiao2026} provide a starting point,
but extensions to nonlinear PDE models remain important.
Particle corrections offer another route
\cite{DL_Cardoso2024,DL_Kelvinius2025}, although their accuracy
and efficiency depend on computable weights, particle degeneracy
and the cost of likelihood evaluations.

\begin{acknowledgement}
The author would like to thank Luca Ratti for giving him the opportunity to present the work at IPMS 2026. This work is supported by the Research Council of Finland (Flagship of Advanced Mathematics for Sensing, Imaging and Modelling, FAME) grant nos. 359186 and 359183. Funding from the Magnus Ehrnrooth Foundation is acknowledged.
\end{acknowledgement}


\end{document}